\documentclass[conference]{IEEEtran}
\usepackage{graphicx}

\usepackage{algorithm}
\usepackage{algorithmic}

\ifCLASSINFOpdf
\else

\fi

\begin{document}
\title{A constrained recursion algorithm for batch normalization of tree-sturctured LSTM}
\author{\IEEEauthorblockN{Ruo Ando}
\IEEEauthorblockA{
National Institute of Informatics\\
2-1-2 Hitotsubashi, Chiyoda-ku, Tokyo 101-8430 Japan
}
\IEEEauthorblockN{Yoshiyasu Takefuji}
\IEEEauthorblockA{
Keio University\\
5322 Endo Fujisawa, Kanagawa 252-0882 Japan
}
}

\maketitle

\begin{abstract}
Tree-structured LSTM is promising way to consider long-distance interaction over hierarchies.
However, there have been few research efforts on the hyperparameter tuning of the construction and traversal of tree-structured LSTM.
To name a few, hyperparamters such as the interval of state initialization, the number of batches for normalization 
have been left unexplored specifically in applying batch normalization for reducing training cost and parallelization.
In this paper, we propose a novel recursive algorithm for traversing batch normalized tree-structured LSTM.
In proposal method, we impose the constraint on the recursion algorithm for the depth-first search of binary tree representation of LSTM
for which batch normalization is applied. 
With our constrained recursion, we can control the hyperparameter in the traversal of several tree-structured LSTMs which is generated in the process of batch normalization. 
The tree traversal is divided into two steps.
At first stage, the width-first search over models is applied for discover the start point of the latest tree-structured LSTM block. 
Then, the depth-first search is run to traverse tree-structured LSTM.
Proposed method enables us to explore the optimized selection of hyperparameters of recursive neural network implementation 
by changing the constraints of our recursion algorithm.
In experiment, we measure and plot the validation loss and computing time with changing the length of internal of state initialization 
of tree-structured LSTM.
It has been turned out that proposal method is effective for hyperparameter tuning such as the number of batches and 
length of interval of state initialization of tree-structured LSTM.
\end{abstract}

\IEEEpeerreviewmaketitle

\section{Introduction}

\subsection{Deep recurrent neural network}

For coping with sequential data which have potentially long-term dependencies, Recurrent Neural Network (RNN) is powerful models.
Recently, the trends of big data technologies, RNN have received renewed attention due to its recent contributions in various domains 
such as speech recognition \cite{alan}, machine translation \cite{Ilya} \cite{Dzmitry} and language modelling
\cite{Tomas}. 
Although recent advances in training RNN, there have been two problems: vanishing gradient and scalability.
First, RNN suffer the difficulty of training by gradient-based optimization procedures. 
Second, we have still fundamental challenge for capturing long-term dependencies in sequences. 
Many proposals leveraging backpropagation is difficult to scale to long-term dependencies. 

As the solution about the difficulties of coping with the long sequences, Long Short Term Memories (LSTM) have been proposed for providing 
the resilience to gradient problems. 
Although there have been many successes by adopting 
chain-structured LSTM, many other important domains are inherently associated with input structures which are more complicated than input sequence itself. 
For example, it is pointed out that sentences in natural languages are 
believed to be carried by not simply a linear sequence of words; 
instead, semantics and its meaning is thought to be nonlinear structures. 
Zhu et al. \cite{zhu} propose a new method for adopting memory blocks in recursive structures.
It is called as S-LSTM of which model utilize the structures and performs better than 
chain-structured LSTM ignoring such priori structures.

Another promising technique for long-term dependencies is batch normalization.
Batch normalization standardizes the inputs to a layer for each mini-batch in training deep neural networks. 
By doing this, the number of training epochs is dramatically reduced with the effect of stabilizing the learning process. 
In detail, batch normalization reduces the amount of covariance shift where the hidden unit values shift around
by normalizing the input layer by adjusting and scaling the activations.

\subsection{Hyper-parameter tuning}
Yet another fundamental challenge of deep neural networks is hyper-parameter tuning. 
Hyper-parameter controls many aspects of the algorithm's behavior, which results in that manual settings 
of hyper-parameter affect the time and memory cost of running the program.
Consequently, automatic hyper-parameter selection can drastically reduce the cost about designing the deep neural network 
algorithm.

Recently, competitive results have been showed by choosing appropriate hyper-parameters for the neural network
to long sequences instead of using a generalized one \cite{Bergstra} \cite{Jozefowicz}.
These hyperparameters define the number of hidden unit per layer, the number of layer, the choice of activation function, 
the kernel size, the length of interval of epochs for batch normalization, etc.
More sophisticated automated hyper-parameter methods are proved to be competitive in selecting deep neural network
settings in comparison to those provided by human experts. 
Unfortunately, automatic hyper-parameter selection is still not generally adopted due to two points.
First, current proposals are not efficient in searching in a high-dimensional hyperparameter space. 
Second, hyper-parameter tuning can usually be an expensive process. 

In this paper, we propose a constrained recursion algorithm for batch normalization of tree-structured LSTM. 
The constrained recursion of our method provides good visibility for selecting hyper-parameters about long-term dependencies. 
Proposal method can evaluate some hyper-parameters such as the number of LSTM module and the state reset interval of each module
with reasonable computing time.
Also, proposal method enables us to avoid the highly cost learning such as weights optimization process.

\section{Methogology}

\subsection{Truncated Backpropagation Through Time}

Backpropagation Through Time, or BPTT, is actually a specific application of back propagation in neural networks applied to sequence data like a time series.
A recurrent neural network is shown one input each time step and predicts one output.
Conceptually, BPTT works by unrolling all input time steps as shown in Figure 1. 
Each time step has one input time step, one copy of the network $s_t$ , and one output $o_t$. 
Errors are then calculated and accumulated for each time step with $w$. 
Figure 1 has outputs at each time step.
The network is rolled back up and the weights are updated. 
BPTT would be impractical in online manner because its memory footprint grows linearly with time.

\begin{figure}
\centering
\includegraphics[scale=0.3]{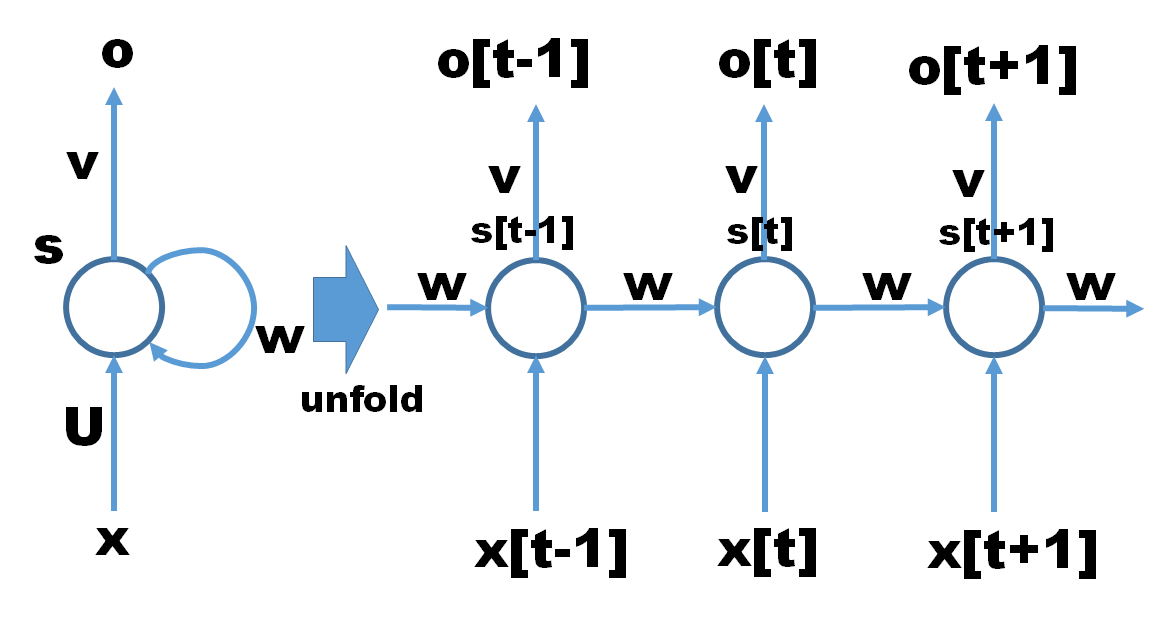}
\caption{Back Propagation Through Time}
\label{fig:verticalcell}
\end{figure}

Truncated Backpropagation Through Time (TBPTT) which is online version of BPTT is proposed in \cite{ronald}.
TBPTT works analogously to BPTT, but the sequence is processed one time step at a time and periodically. 
the BPTT update is performed back for a fixed number of time steps.
In \cite{ronald}, the accumulation stops after a fixed number of time steps.
Truncated BPTT performs well if the truncated chains are effective 
to learn the target recursive functions.

\subsection{LSTM}
Long short-term memory (LSTM) \cite{hochreiter} is a family of recurrent neural network.
Like other recurrent neural networks, LSTM has feedback connections. 
Concerning the memory cell itself, it is controlled with a forget gate, which can reset the memory.
unit with a sigmoid function. In detail, given a sequence data $ {x_1, ..., x_T } $we have the gate definition as follows:

\begin{eqnarray}
f_{t} = \sigma ( W_{fh}h_{t-1} + W_{fx}x_{t} + P_{f} * c_{t-1} * b_{f} ) \\
i_{t} = \sigma ( W_{i}x_{t} + U_{i}h_{t-1} + P_{i} * c_{t-1} * b_{i} ) \\
g_{t} = tanh ( W_{g}x_{t} + U_{g}h_{t-1} + b_{g} ) \\
c_{t} = i_{t} \Theta g_{t} + f_{t} \Theta c_{t-1} \\
o_{t} = \sigma ( W_{o}x_{t} + U_{o}h_{t-1} + P_{o} * c_{t} + b_{o} ) \\ 
h_{t} = o_{t} \Theta tanh (c_{t}) 
\end{eqnarray}
where $ f_t $ is forget gate, $ i_t $ input gate, ot output gate and $ g_t $input modulation gate. 
Particularly $P_f, P_i P_o$ indicates the peephole weights for the forget gate.
The peephole connections introduced in \cite{Gers} enables the LSTM cell to inspect its current internal states.
Then, the backpropagation of the LSTM at the current time step t is as follows:
\begin{eqnarray}
\delta o_{t} = tanh ( c_{t}) \delta h_{t} \\ 
\delta c_{t} = (1 - tanh(c_{t})^2) o_{t} \delta h_{t} \\
\delta f_{t} = c_{t-1} \delta c_{t} \\
\delta c_{t-1} = f_{t} \theta \delta c_{t} \\ 
\delta i_{t} = g_{t} \delta c_{t} \\
\delta g_{t} = i_{t} \delta c_{t} 
\end{eqnarray}

\subsection{Batch normalization}

Another exciting deep learning technique is batch normalization which reparameterizes the model for introducing both additive and multiplicative
noise on the hidden layer at the training rate. That is, batch normalization is a method for adaptive reparameterization driven by the difficulty of training deep neural network.
Originally, batch normalization is introduced by \cite{Sergey}. In \cite{Sergey},  a method for addressing the vanishing/exploding gradients problems is proposed.
Specifically, this is described by the Internal Covariate Shift problem where the distribution of each layer's inputs changes during training.
The whitening each layer of the network for the reduction of internal covariance shift takes too computationally demanding.
Batch normalization takes advantage in approximating the whitening with the statistics of the current mini batch.
Given a mini-batch x, the sample mean and sample variance of each feature k can be calculated. 

\[
\overline{x_k} = \frac{1}{m} \sum_{i=1}^m x_i, k,
\]

\[
\sigma_k^2 = \frac{1}{m} \sum_{i=1}^m ( x_i, k - \overline{x_k} ) ^2
\]

where m is the length of the mini-batch. Using $overline{x_k}$ and $\sigma_k^2$, we can standardize each feature k as
follows:

\[
\hat{x_k} = \frac{x_k -\overline{x_k} }{\sqrt{\sigma_k^2 + \epsilon}}
\]

where $\epsilon$ is a small positive constant to keep numerical stability.
Although the main purpose of batch normalization is to simplify optimization, the noise can have a regularizing effect and batch normalization sometimes makes dropout unnecessary. 
%Batch normalization can have a dramatic effect on optimization performance, especially for convolutional networks and networks with sigmoidal nonlinearities.

\section{Proposal method}

\subsection{Tree-structured LSTM}

The Tree-LSTM is a generalization of long short-term memory (LSTM) networks to tree-structured network topologies, introduced in \cite{zhu}.
The core idea is to introduce syntactic information for language tasks by extending the chain-structured LSTM to a tree-structured LSTM. 
%The dependency tree and constituency tree techniques are leveraged to obtain a ``latent tree''.
Figure 2 shows the comparison of two kinds of LSTM network structures.
The upper side of Figure 2 shows a chain-structured LSTM network.
The lower side of Figure 2 depicts a tree-structured LSTM network with arbitrary branching factor.
Tree structured LSTM performs well when the networks need to combine words and phrases in natural language processing \cite{tai}. 

\begin{figure}
\centering
\includegraphics[scale=0.5]{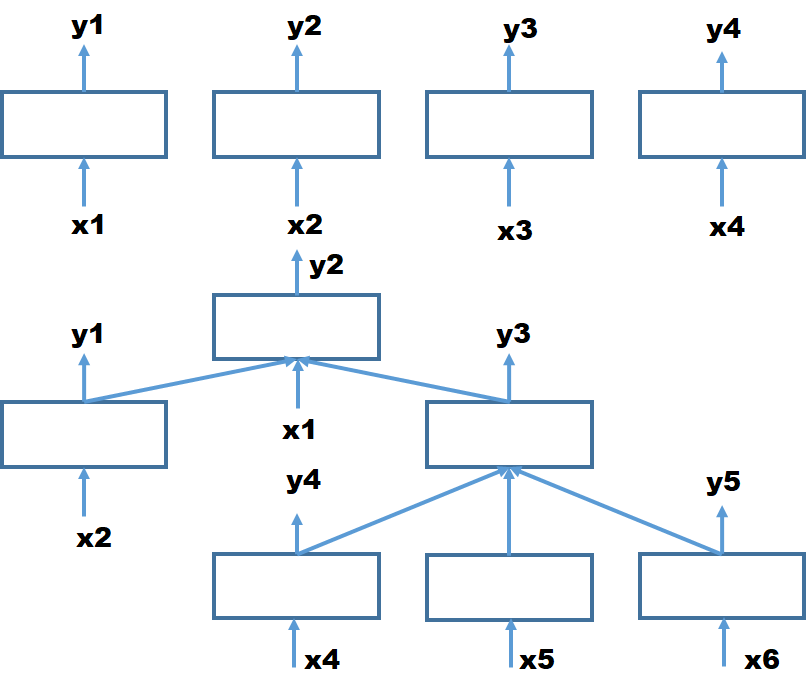}
\caption{Tree structured LSTM }
%\label{fig:verticalcell}
\end{figure}

\begin{figure*}[t]
\centering
\includegraphics[scale=0.5]{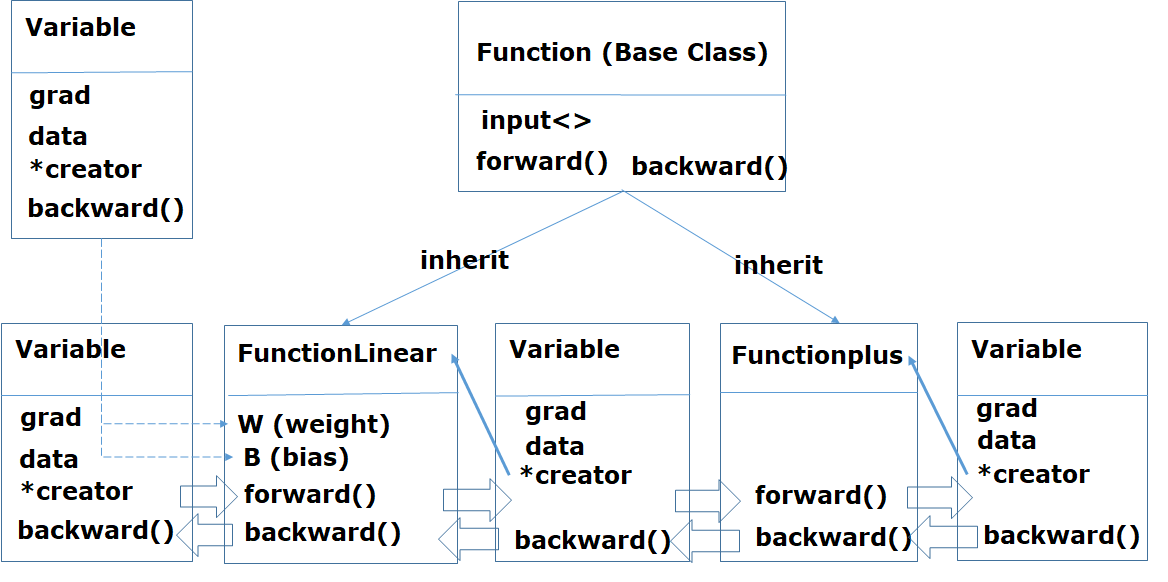}
\caption{Reverse-mode autodiff of linear activation unit}
\label{fig:verticalcell}
\end{figure*}

\subsection{Reverse-mode diff}

Autodiff (Automatic differentiation) is the fundamental technique upon which most deep learning frameworks is based. 
Autodiff is one of the gradient based techniques which deep learning models use and autodiff enables makes it easy to calculate gradients, even from enormous and complicated models.
Reverse-mode autodiff which is implemented in Tensorflow \cite{tensorflow} is a very powerful and accurate technique, specially in the case that there are many inputs and few outputs.
At first phase of reverse-mode diff, program goes through the graph in the forward direction from the input to the output to calculate the value of each mode. 
At second phase program goes through the reverse direction in turn to compute all the partial derivatives.
Besides, reverse-mode diff can deal with functions defined by arbitrary code. 

Figure 3 depicts our implementation of reverse-mode autodiff of linear activation. 
In artificial neural networks, the activation function of a node defines the output of that node given an input or set of inputs. 
Input-output model is defined as follows:

\[
f(x) = \psi * (\sum_{i=0}^n w_i * x_i + b) 
\]

Here, $ \psi $ is activation function such as Tanh and RELU. 
Class FunctionLinear implements the function of $ \sum_{i=0}^n w_i * x_i + b $.
The notation of *creator is the pointer to the function which generates its variable. 
For example, FunctionLinear outputs $ r $ which is equal to $ \sum_{i=0}^n w_i * x_i + b $ and is passed to FunctionTanh. 
The creator of variable $ r $ is FunctionLinear.

Figure 3 also illustrates the detailed implementation of reverse-mode autodiff of tree-structured LSTM.
Inheritance lets us define classes which model relationships among types, sharing what is common and specializing
only that which is inherently different. members defined by the base class are inherited by its derived classes. 
The derived class can use, without change, those operations that do not depend on the specifics of the 
derived type. It can redefine those member functions which do depend on its type,
specializing the function to take into account the peculiarities of the derived type. 
Finally, a derived class may define additional members beyond those it inherits from its base class.

\subsection{Recursion and binary tree of LSTM}

Recursion is a fundamental process in different modalities. 
Also, recursion is core technique for traversing binary tree. Binary tree is a tree whose elements have at most 2 children.
Since each element in a binary tree can have only 2 children, we typically name them the left and right child.
Recursion is a fundamental process associated with many problems - a recursive process and hierarchical structure so formed are common in different modalities.
The forward computation of a S-LSTM memory block is specified in the following equations.

\begin{eqnarray}
%\begin{equation}
%\begin{split}
i_{t}=\sigma(W_{hi}^{L}h_{t-1}^{L}+ W_{hi}^{R}h_{t-1}^{R}) \nonumber \\
+ W_{ci}^{L}c_{t-1}^{L}+ W_{ci}^{R}c_{t-1}{R} + b_{i}) \\ 
f_{t}^{L}=\sigma(W_{fl}^{L}h_{t-1}^{L}+W_{hfl}^{R}h_{t-1}^{R} \nonumber \\ 
+W_{cfl}^{L}c_{t-1}^{L}+W_{cfl}^{R}c_{t-1}{R} + b_{fl}) \\
f_{t}^{R}=\sigma (W_{ft}^{L}h_{t-1}^{L}+W_{hfr}^{R}h_{t-1}^{R} \nonumber \\
+W_{cfr}^{L}c_{t-1}^{L}+W_{cfr}^{R}c_{t-1}{R}+b_{fr}) \\
x_{t}=W_{hx}^{L}h_{t-1}^{L} \nonumber \\
+W_{hx}^{R}h_{t-1}^{R}+b_{x} ) \\
c_{t}=f_{t}^{L}*c_{t-1}^{L}+f_{t}^{R}*i_{t}*tanh (x_{t}) \\
o_{t}=\sigma(W_{ho}^{L} h_{t-1}^{L} \nonumber \\
+W_{ho}^{R} h_{t-1}^{R}+W_{co}c_{t}+b_{o})\\
h_{t}=o_{t}*tanh(c_{t}) 
%\end{flushleft}
%\end{split}
%\end{equation}
\end{eqnarray}
%\end{flushleft}

where $ \sigma $ is the element-wise logistic function used to confine
the gating signals to be in the range of [0, 1]. $ f_{L} $
and $ f_{R} $ are the left and right forget gate, respectively; b
is bias and W is network weight matrices; the sign * is a Hadamard product, i.e., element-wise product.

More importantly, equation (13)-(19) consist of binary operator.
Therefore, these equations can be represented as binary tree. 
In general, binary tree is a data structure whose elements have at most 2 children is called a binary tree. 
Figure 4 depicts our binary tree representation of LSTM.
Since each element in a binary tree can have only 1 child or 2 children, we typically call them the left and right child. 
The topmost node in the tree is named as the root which is h in this case.

\begin{figure}
\centering
\includegraphics[scale=0.45]{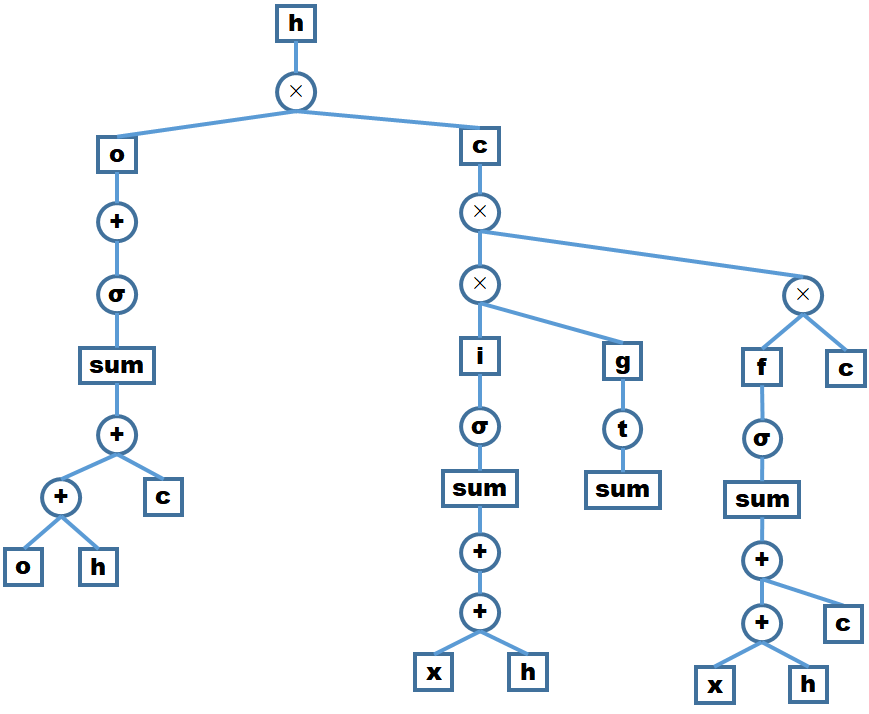}
\caption{Binary tree of LSTM }
\label{fig:verticalcell}
\end{figure}

\begin{algorithm}
\caption{recursive function}
\label{alg1}
\begin{algorithmic}[1]
\renewcommand{\algorithmicrequire}{\textbf{Input:}}
\renewcommand{\algorithmicensure}{\textbf{Output:}}

\STATE $ variable \rightarrow generator \rightarrow backward(grad) $
\WHILE{$ i \leq variable \rightarrow generator.inputs\_size() $}
% \STATE form k clusters by assigning each points in X to its nearest center
\STATE $ nv = variable \rightarrow generator.inputs() $
\IF{$ nv = isGetGrad $}
\STATE $ this \rightarrow backward(nv.get())$
\ENDIF
\ENDWHILE
\end{algorithmic}
\end{algorithm}

\subsection{Constrained model traversal}

Our constrained recursion algorithm is based on the division of all learning process into model and graph.
Figure 5 depicts our algorithm.
Model block consists of Mean Squared Error (MSE) \textcircled{1}\textcircled{4}\textcircled{7}, Tanh \textcircled{2}\textcircled{5}\textcircled{8}and LSTM tree \textcircled{3}\textcircled{6}\textcircled{9}.
Each LSTM tree node has LSTM graph\textcircled{10}\textcircled{11}\textcircled{12}. 

Constrained model traversal is designed to reach the latest graph generated in \textcircled{10} and start the backpropagation. 
To do this, the program should search model block in depth-first search manner. 
Depth-first search is the extension of basic tree traversal, which adopts recursive graph traversal as shown at the line 5 of Algorithm 1.
Recursive graph traversal can systematically visit every node in a graph is a direct generalization of the tree-traversal methods. 
However, in the case of Figure 5, without any constraint, the traversal path of the program can be misguided to the path of \textcircled{1}, \textcircled{2}, \textcircled{3} and reaching \textcircled{12}.
To prevent this wrong path, at the node of LSTM tree \textcircled{3}\textcircled{6}\textcircled{9}, adopt Algorithm 2. 
At line 1-3, the program checks whether the variable under traversing is the latest generated graph (which is \textcircled{10}).

\begin{algorithm}
\caption{Constrained model traversal}
\label{alg1}
\begin{algorithmic}[1]
\IF{$v \rightarrow last_{opt} \neq NULL \land \rightarrow opt = *v \rightarrow last_{opt} $ }
\STATE $ *v \rightarrow is\_last\_backward = true $
\ENDIF
\IF{$if (v \rightarrow is\_last\_backward \neq NULL \land *v \rightarrow is\_last\_backward = false $}
\STATE $ return $
\ENDIF
\STATE $ back\_propagation() $
\end{algorithmic}
\end{algorithm}

\begin{figure*}
\centering
\includegraphics[scale=0.5]{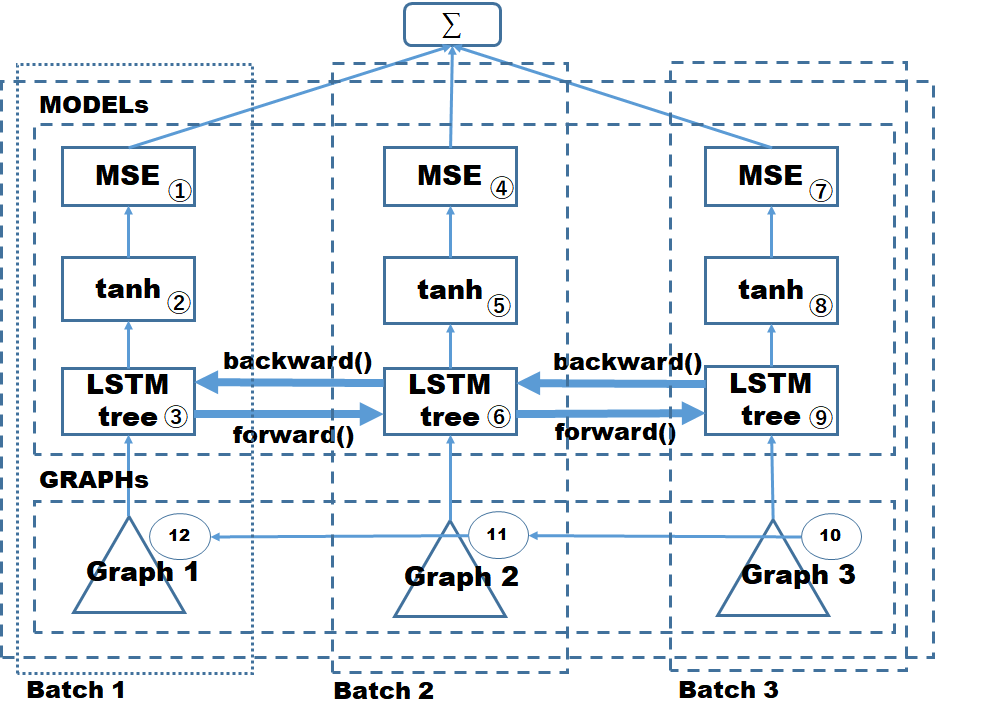}
\caption{Contrained model traversal }
\label{fig:verticalcell}
\end{figure*}

\subsection{Constrained graph traversal}

At first phase of LSTM, the current input layer x(t) and the previous short-term state h(t-1) are fed to four differently connected layers. 
Four connected layers are g (main gate), i (input gate), f (forget gate) and o (output gate). 
Here, let us call the multiple connection as branch as shown in Figure 5.
Without any constrain, the propagation between x(t) and these four layers are duplicated. 
We have solution for the problem of this duplication with Algorithm 3.

In Figure 6, variable X has four branches reaching to g, i, f and o. 
In forward propagation phase, variable x has been allocated four times as x(1), x(2), x(3) and x(4).
Without any constraint, the backpropagation from x will be caused four times to fx, gx, ix and ox.
To stop this duplicated propagation, we set the condition at line 4 of Algorithm 3.

\begin{figure}
\centering
\includegraphics[scale=0.5]{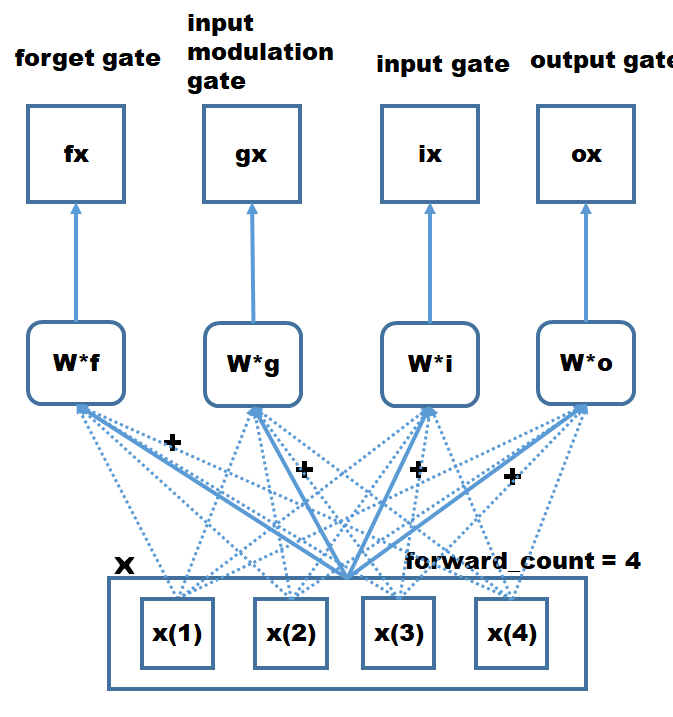}
\caption{Multiple connection (four branches)}
\label{fig:verticalcell}
\end{figure}

\begin{algorithm}
\caption{Constrained graph traversal}
\label{alg1}
\begin{algorithmic}[1]
\IF{$v iv \rightarrow forward\_count > 0 $ }
\STATE $ v \rightarrow forward\_count-- $
\ENDIF
\IF{$ v \rightarrow forward_count \neq 0 $}
\STATE $ reutrn $
\ENDIF
\STATE $ back\_propagation() $
\end{algorithmic}
\end{algorithm}

\begin{figure*}
\centering
\includegraphics[scale=0.5]{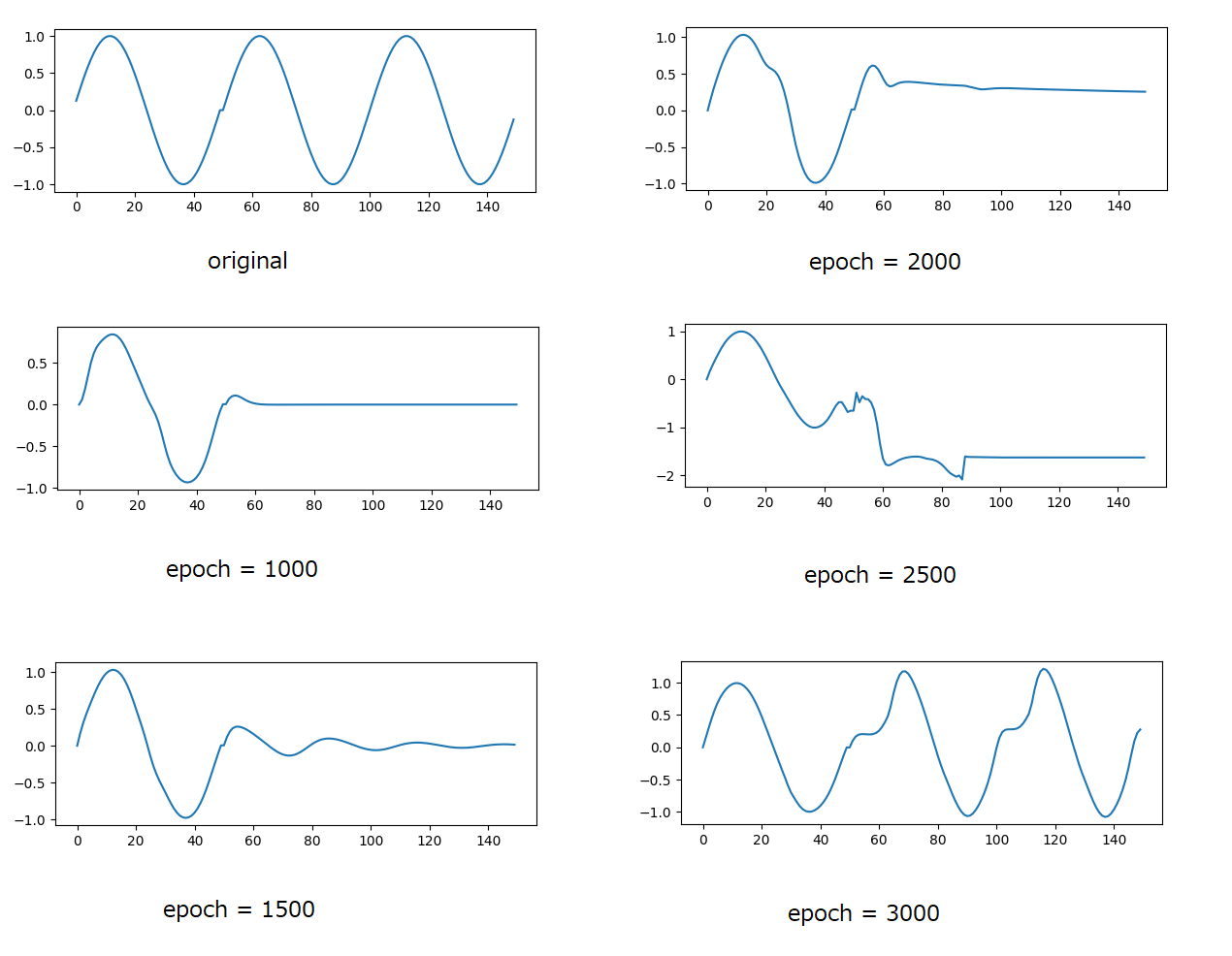}
\caption{Prediction plots}
\label{fig:verticalcell}
\end{figure*}

\section{Experiment}

In this section we describe the experimental results of the training and generating sine wave. 
In experiment, we use workstation with Intel(R) Xeon(R) CPU E5-2620 v4 (2.10GHz) and 251G RAM.

Figure 7 shows the comparison of predicted sequence generated by proposal LSTM implementation. 
The x axis is training epochs. Y axis shows the input and predicted value of sine wave. 
Predicted sequences are generated with epoch 1000, 1500, 2000, 2500 and 3000.
The result reveals the significant improvement between epoch 2500 and 3000.
The best acceleration with minimum loss was achieved when epoch=3000.

Figure 8 depicts the validation loss of proposed LSTM implementation with epoch 1000.
We compare the results with the interval length of model state initialization of 5, 10 and 15.
The interval length of model reset (state initialization) has a significant effect on the average loss.
Here, our definition of efficiency is the rate of change in loss value with respect to training steps.
And we denote $ intvl $ as the length of interval of model state initialization. 
The average loss improves as we increase the amount of look-ahead and plateaus after 200 epochs. 
The cases of $ intvl = 5, 10, 15 $ converged within around 600 training steps, but the the case of $ intvl=5 $
delivered a lower convergence speed with smaller training loss.

Figure 9 shows the comparison of elapsed timer per one step with epoch=1000.
As the learning process is going forward, the elapsed time taken per one iteration is increasing. 
In this point of view, the longer interval time is effective. 
Considering reaching the plateau around the 800 epochs, it can be concluded that the case of $ intvl=10 $ 
is feasible with lesser than about 20\% elapsed time.

These results of Figure 8 and 9 have implications that we can validate the effectiveness of changing $intvl$ length for achieving lower average loss 
while our program keeps the reasonable computing resource utilization.

\begin{figure}
\centering
\includegraphics[scale=0.5]{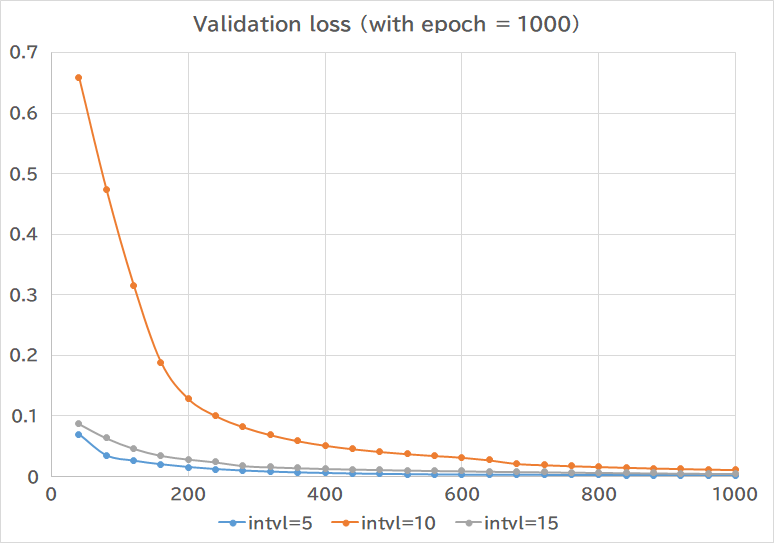}
\caption{Validation loss with epoch 1000}
\label{fig:verticalcell}
\end{figure}

\begin{figure}
\centering
\includegraphics[scale=0.5]{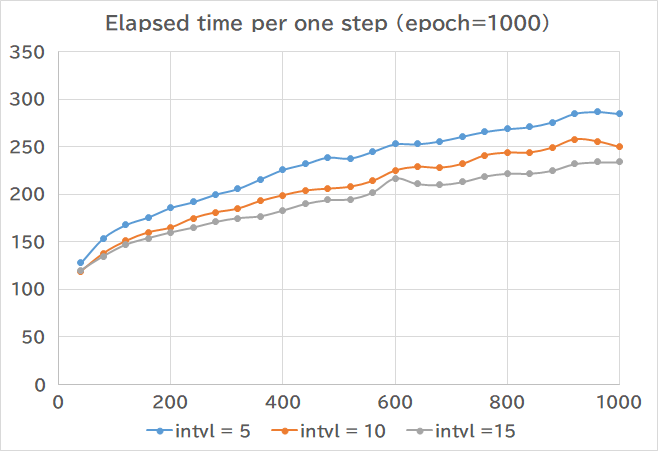}
\caption{Elapsed time per iteration with epoch=1000}
\label{fig:verticalcell}
\end{figure}

\section{Related work}

Recurrent neural networks \cite{Rumelhart}, or RNNs are feedforward neural networks for processing sequential data 
by extending with incorporating edges that span adjacent time steps. 
In general, RNNs suffer the difficulty about training by gradient-based optimization procedures.
Local numerical optimization includes stochastic gradient descent or second order methods, which causes the 
exploding and the vanishing gradient problems\cite{yoshua}\cite{john}\cite{razvan}.
Werbos et al. \cite{werbos1} propose the backpropagation through time (BPTT) which is training algorithm
for RNN. BPTT is derived from the popular backpropagation training algorithm used
in MLPN training \cite{werbos2}.
Derivatives of errors are computed with backpropagation over structures \cite{goller}.

Recursive neural networks are yet another representation of the generalization of recurrent networks 
with a different kind of computation graph. The computation graph adopted in recursive neural networks
is a deep tree, instead of the chain-like structure of RNNs.
Pollack \cite{pollack} proposes recursive neural networks. Bottou \cite{bottou} discuss the potential use 
of recursive neural network in learning to reason. 
In \cite{socher1} and \cite{socher2}, recursive neural networks are more effective in performing 
on different problems such as semantic analysis in natural language processing and image segmentation.

There is a long line of research efforts on extending the standard LSTM \cite{hochreiter} in order
to adopt more complex structures.
Tai et al \cite{tai} and Zhu et al. \cite{zhu} extended chain-like structured LSTMs to tree-structured LSTMs by adopting
branching factors. 
They demonstrated that such extensions outperform competitive LSTM baselines on several
tasks such as semantic relatedness prediction and sentiment classification.
Furthermore, Li et al. \cite{li} show the effectiveness of tree-structured LSTM on various
tasks and situations in which tree-like structure is effective.

Truncated BPTT \cite{ronald} is one of the most popular variants of BPTT.
In \cite{ronald}, the accumulation stops after a fixed number of time steps.
Truncated BPTT performs well if the truncated chains are effective 
to learn the target recursive functions.
Saon et al \cite{george} improved the original truncated BPTT 
with batch decoding. In \cite{george}, the number of context frames
in batch decoding is equals to the number of unrolled steps before truncation.

Practical recurrent networks are combined of BPTT, batch decoding and consecutive prediction \cite{hasim} \cite{kai} for speeding up training. 
Besides, the hidden vectors are sometimes cached in \cite{george}.
These techniques sometimes cause a situation mismatching between training and testing. 
So far, this mismatch has not been addressed. However, in \cite{kanda}, 
the distinction between online and batch decoding under running BPTT is explored.

\section{Discussion}
As we noted in the previous section, recursive neural networks use a different kind of computational graph for adopting yet another generalization of recurrent networks. Originally, Pollack \cite{pollack} introduces recursive neural networks. Bottou \cite{bottou} shows the potential use for learning to reason. So far, what deep learning achieved is the ability to map set X to set Y using continuous geometric transformation with given large amounts of human-annotated data.
This straightforward geometric morphing from space X to space Y of deep-learning models do is called as local generalization.
In the future, a necessary transformational development in the field of machine learning is move away from local generalization with purely pattern recognition towards model capable of the extreme generalization with abstraction and reasoning.
As we already know, a likely appropriate substrate in various situations is computer program.
Takefuji \cite{Takefuji} points out that the importance of modularity and abstraction for the progress of the open source software such as Keras \cite{keras} , Chainer\cite{chainer} , and Pytorch\cite{pytorch} . 
With the progress of software engineering, modularity and abstraction in software development will becomes a fundamental sense for achieving higher reusability with being more resilient against input/output/process interactions.

Batch normalization applied in this paper is not actually an optimization algorithm but a technique of adaptive representation. 
As illustrated in section III, with batch normalization, each tree-structured LSTM block is modularized and linked to the node of summation of loss (see Figure 5). 
Our network with batch normalization adopted can be described as shallow learning network and similar to functional link network.
Pao and Takefuji \cite{Pao} proposes the functional link model which eliminates all layers between input and output by
using single step of processing is one way to avoid the nonlinear
learning. One benefit of functional link neural network is flexibility when the learning time is based on the numerous processing elements necessary for computing. 

\section{Conclusion}
In this paper we have proposed the novel constrained recursion algorithm for the construction and traversal of tree structured LSTM.
Tree-structued LSTM is promising way to consider long-distance interaction over hierarchies.
However, there have been few research efforts on the hyperparameter tuning in coping with tree-structured LSTM.
Specifically, the hyperparamter of the interval of state initialization have been left unexplored in applying batch normalization.
With our constrained recursion, we can control the hyperparameter in the traversal of several tree-structured LSTMs which is generated in the process of batch normalization. 
The tree traversal is divided into two steps.
At first phase, the width-first search over models is applied for discover the start point of the latest tree-structured LSTM block. 
The second phase is the graph traversal of each tree-structured LSTM by the depth-first search.
Proposed method makes it possible to validate the optimized selection of hyperparameters of recursive neural network implementation 
by changing the constraints of our recursion algorithm.
In experiment, we measure and plot the validation loss and computing time with changing the length of internal of state initialization 
of tree-structured LSTM. The average loss improves as we increase the amount of look-ahead and plateaus after 200 epochs. 
The cases of $ intvl =5, 10 and 15 $ converged within around 600 training steps, but the case of $ intvl=5 $
delivered a lower convergence speed with smaller training loss.
It has been turned out that proposal method is effective for hyperparameter tuning of the interval of tree-structured LSTM.
For future work, we are going to figure out the dynamic state initialization by using our constrained recursion algorithm.

% that's all folks
\end{document}